\documentclass[conference]{IEEEtran}
\IEEEoverridecommandlockouts
\usepackage{cite}
\usepackage{amsmath,amssymb,amsfonts}
\usepackage{algorithmic}
\usepackage{graphicx}
\usepackage{textcomp}
\usepackage{xcolor}
\usepackage{hyperref}
\usepackage{multirow}
\def\BibTeX{{\rm B\kern-.05em{\sc i\kern-.025em b}\kern-.08em
    T\kern-.1667em\lower.7ex\hbox{E}\kern-.125emX}}
\newtheorem{definition}{Definition}
\DeclareMathOperator*{\argmax}{arg\,max}
\begin{document}

\title{Representation Selection via Cross-Model \\ Agreement using Canonical Correlation Analysis}

\author{\IEEEauthorblockN{Dylan B. Lewis}
\IEEEauthorblockA{
\textit{University of Tennessee}\\
Knoxville, USA \\
dlewis37@vols.utk.edu}
\and
\IEEEauthorblockN{Jens Gregor}
\IEEEauthorblockA{
\textit{University of Tennessee}\\
Knoxville, USA \\
jgregor@utk.edu}
\and
\IEEEauthorblockN{Hector Santos-Villalobos}
\IEEEauthorblockA{
\textit{University of Tennessee}\\
Knoxville, USA \\
hsantosv@utk.edu}
}

\maketitle

\begin{abstract}

Modern vision pipelines increasingly rely on pretrained image encoders whose representations are reused across tasks and models, yet these representations are often overcomplete and model-specific. We propose a simple, training-free method to improve the efficiency of image representations via a post-hoc canonical correlation analysis (CCA) operator. By leveraging the shared structure between representations produced by two pre-trained image encoders, our method finds linear projections that serve as a principled form of representation selection and dimensionality reduction, retaining shared semantic content while discarding redundant dimensions. Unlike standard dimensionality reduction techniques such as PCA, which operate on a single embedding space, our approach leverages cross-model agreement to guide representation distillation and refinement. The technique allows representations to be reduced by more than 75\% in dimensionality with improved downstream performance, or enhanced at fixed dimensionality via post-hoc representation transfer from larger or fine-tuned models. Empirical results on ImageNet-1k, CIFAR-100, MNIST, and additional benchmarks show consistent improvements over both baseline and PCA-projected representations, with accuracy gains of up to 12.6\%.

\end{abstract}

\section{Introduction}
\label{sec:intro}

Modern vision systems increasingly rely on large pretrained encoders whose representations are reused across tasks, datasets, and downstream pipelines \cite{dosovitskiy2021imageworth16x16words, caron2021emergingpropertiesselfsupervisedvision, he2021maskedautoencodersscalablevision, assran2023selfsupervisedlearningimagesjointembedding}. While these representations are highly expressive, they are often overcomplete and model-specific, and there is limited guidance on which dimensions capture task-relevant semantic structure across models as opposed to incidental or redundant variation. Common dimensionality reduction techniques such as principal component analysis (PCA) operate on single-model statistics \cite{Hotelling_1933}, making it difficult to distinguish generally useful structure from artifacts specific to a particular training objective or architecture.

At the same time, it has become increasingly common to work with multiple pretrained models—often trained using different self-supervised objectives or fine-tuned to different extents—without retraining them jointly~\cite{radford2021learningtransferablevisualmodels,ilharco_gabriel_2021_5143773}. Prior work has explored representation alignment and comparison across networks primarily as an analysis tool rather than as a mechanism for modifying representations themselves \cite{raghu2017svccasingularvectorcanonical, morcos2018insightsrepresentationalsimilarityneural}. This raises a natural question: can agreement between independently trained representations be leveraged directly to identify and reuse shared semantic structure post-hoc, without additional optimization or supervision

\begin{figure}[t]
    \centering
    \includegraphics[width=\columnwidth]{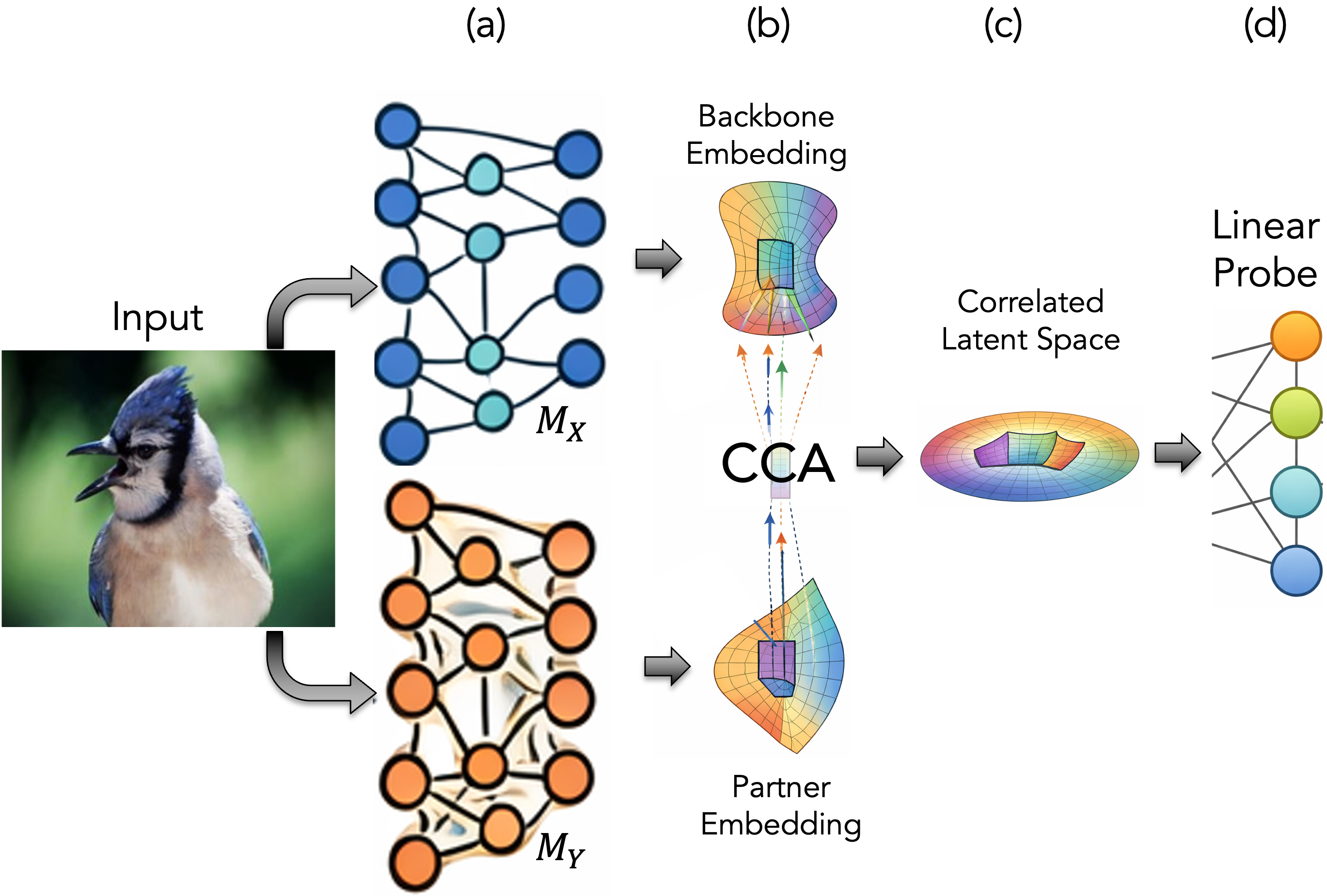}
    \caption{CCA cross-model alignment pipeline. (a) A backbone encoder $M_X$ and a partner encoder $M_Y$ produce baseline embeddings from the same input image. (b) CCA projects both embeddings into a shared, maximally correlated subspace. (c) The aligned space enables dimensionality reduction while transferring shared structure across models. (d) The resulting representations are evaluated on a downstream classification task. }
    \label{fig:pipeline}
\end{figure}

In this paper, we re-purpose canonical correlation analysis (CCA) as a tool for cross-model representation selection, identifying directions in two embedding spaces that are mutually predictable and therefore shared across independently trained encoders. This perspective re-frames CCA from an analysis technique into an operator for aligning and filtering representations based on cross-model agreement. We emphasize that our goal is not to compete directly with state-of-the-art methods specialized for distillation, dimensionality reduction, or representation learning, which typically rely on task-specific training objectives. Instead, we study a complementary setting in which representations are aligned and refined post-hoc, without retraining, supervision, or architectural modification. \hyperref[fig:pipeline]{Figure~\ref{fig:pipeline}} illustrates the data and process flow for our method. 

We show the effectiveness of our method on a variety of image classification datasets, including ImageNet-1k, CIFAR-100, and MNIST. The main findings of this work are the following: 
\begin{itemize}
    \item \textbf{Post-hoc representation selection via agreement:} We repurpose canonical correlation analysis as a post-hoc representation selection operator, showing that cross-model agreement identifies semantically useful subspaces shared across independently trained image encoders $M_X$ and $M_Y$.
	\item \textbf{Agreement-based selection outperforms variance-based reduction:} We empirically demonstrate that CCA-based projections consistently improve linear evaluation performance relative to both unprojected baseline representations and PCA-based dimensionality reduction, indicating that cross-model agreement is a stronger inductive bias than single-model variance.
	\item \textbf{Post-hoc transfer of fine-tuning signal without training:} We show that when two encoders share pre-training but only one is fine-tuned on a downstream task, CCA-based alignment enables post-hoc transfer of task-relevant structure, improving performance for both fine-tuned and non-fine-tuned models without additional optimization.
	\item \textbf{Deterministic, model-driven dimensionality selection:} We demonstrate that CCA induces a deterministic target dimensionality based on model pair geometry, enabling substantial compression without dataset-specific tuning.
	\item \textbf{Robustness across datasets and identification of failure modes:} We validate the method across multiple vision benchmarks and characterize class imbalance as a structured failure mode, providing guidance on when CCA-based alignment is reliable.
\end{itemize}

This work provides a PyTorch \cite{paszke2019pytorchimperativestylehighperformance} implementation of the CCA computation from \cite{raghu2017svccasingularvectorcanonical, morcos2018insightsrepresentationalsimilarityneural} which
makes the CCA computation faster by utilizing GPUs.

\section{Related Work}
\label{sec:rel-works}

\subsection{Canonical Correlation Analysis}
\label{sec:cca}
Canonical correlation analysis (CCA) is a classical technique from multivariate statistics \cite{da6385d2-9c65-3860-bbcd-b821fdff69ff} that 
determines how to linearly project two datasets into a maximally correlated, joint-dimensional latent space by solving a pair of generalized eigenvalue problems.
Generalizations include kernel CCA \cite{10.1162/153244303768966085, 10.5555/1248659.1248673}, which finds nonlinear dependencies between datasets, and multi-set CCA \cite{Nielsen2002-hu}, which extends application to multiple datasets. CCA has been applied to signal processing \cite{Safieddine_Kachenoura_Albera_Birot_Karfoul_Pasnicu_Biraben_Wendling_Senhadji_Merlet_2012}, unlabeled clustering \cite{chaudhuri2009multi}, multi-view learning \cite{wang2016deepmultiviewrepresentationlearning}, and statistical data fusion \cite{Nielsen2002-hu}, serving as a foundational tool for understanding relationships between heterogeneous representations.
Deep CCA \cite{pmlr-v28-andrew13} and related derivatives \cite{benton2017deepgeneralizedcanonicalcorrelation}, \cite{9428614} learn nonlinear relationships between datasets using gradient-based optimization which enables richer cross-view correspondences. However, these methods more closely resemble traditional multi-view and contrastive representation learning methods.

In addition, CCA has been widely used to analyze and compare internal representations learned by neural networks. Methods such as SVCCA \cite{raghu2017svccasingularvectorcanonical} and projection-weighted CCA \cite{morcos2018insightsrepresentationalsimilarityneural} employ CCA to quantify representational similarity across architectures, layers, or training stages, providing insight into how networks evolve during learning. In these works, CCA serves as a diagnostic tool: the learned projections are used to measure similarity but are not applied to modify or deploy representations in downstream tasks. In contrast, our work uses CCA-derived projections directly as representation transformations, enabling post-hoc alignment, selection, and reuse of representations without additional training.

\subsection{Dimensionality Reduction}
Dimensionality reduction is a well-studied problem. Widely used techniques include principal component analysis (PCA)  \cite{Hotelling_1933} and linear discriminant analysis (LDA) \cite{fisher1936lda},
Non-linear variants of PCA \cite{10.1007/BFb0020217} and LDA \cite{NIPS1999_c0d0e461} also exist, but our work utilizes the classical, linear formulations of these methods. 
PCA based dimensionality reduction retains the dimensions that account for most of the variance within a single dataset, while LDA retains the dimensions that capture the most intra-class variance (directions that separate classes the best). In either case, the number of dimensions retained depends on intrinsic properties of the data considered. 
CCA, on the other hand, produces a space with a pre-defined dimensionality. We propose using this property to guide dimensionality reduction for image representations in a more deterministic manner. By using two models, one that yields a lower dimensional representation than the other, the high-dimensional representation can be compressed to the low-dimensional representation, regardless of the data itself.  

\subsection{Representation Learning}
\label{sec:representation-learning}
The practice of pre-training models to produce strong representations of large datasets followed by fine-tuning to perform various tasks on different datasets has become a standard practice in computer vision. Methods such as 
SimCLR \cite{chen2020simpleframeworkcontrastivelearning}, MoCo \cite{he2020momentumcontrastunsupervisedvisual}, BYOL \cite{grill2020bootstraplatentnewapproach}, SimSiam \cite{chen2020exploringsimplesiameserepresentation},
were shown to work very well for the model pre-training portion of the aforementioned pipeline. State-of-the-art methods such as DINO \cite{caron2021emergingpropertiesselfsupervisedvision}, joint-embedding predictive architectures \cite{assran2023selfsupervisedlearningimagesjointembedding}, and masked autoencoders \cite{he2021maskedautoencodersscalablevision} combine principles from contrastive methods and vision transformers to produce high-quality representations of images. 

The core objective of the aforementioned methods is to learn strong visual representations, either by explicitly aligning semantically similar samples in a latent space (e.g., contrastive and DINO-style methods) or by learning representations through reconstruction-based objectives (e.g., masked autoencoders). The representations produced by models trained using these techniques can be classified with high accuracy using a simple linear classification head, described in several of the above works, and this protocol is how we evaluate the representations in our experiments. 

\section{Methodology}
\label{sec:method}

In this work, two types of models are used. The first is a pre-trained, frozen vision transformer that produces high-dimensional representations of images, and the second is a simple linear layer trained to classify these representations. Using a simple linear classifier to evaluate the quality of representations is a widely used technique described in several of the works listed in \hyperref[sec:representation-learning]{Section~\ref{sec:representation-learning}}. The following definitions formally define these types of models and their outputs.

\begin{definition}
    \label{def:img-representation}
    Vector $\Vec{x} \in \mathbb{R}^d$ is a high-dimensional representation of an input image obtained by mean-pooling the output of the last attention block of a vision transformer and applying $\ell_2$ normalization to the resulting vector.
\end{definition}

\begin{definition}
    \label{def:linear-classifier}
    Let $\Vec{y} = W\Vec{x} + \Vec{b}$ denote a linear classifier, where matrix $W$ and vector $\Vec{b}$ are learned weights and biases, and $\argmax(\Vec{y})$ is the predicted class of representation $\Vec{x}$.
\end{definition}

\subsection{Vision Transformer Backbones}

We evaluate our approach using Vision Transformer (ViT) encoders \cite{dosovitskiy2021imageworth16x16words}, which have become a standard backbone for representation learning in computer vision. We consider models of varying capacity and representation dimensionality to study how cross-model alignment behaves across different scales. Unless otherwise noted, all encoders are pretrained using established self-supervised or supervised protocols and remain frozen throughout our experiments.

For experiments using the ImageNet-1k dataset \cite{deng2009imagenet}, we use vision transformers pre-trained on ImageNet-21k and fine-tuned on ImageNet-1k using the method given by \cite{steiner2022trainvitdataaugmentation}.
We obtain checkpoints of parameters for these models from \cite{rw2019timm} to avoid the large computational cost required to train these models from scratch. The four sizes of vision transformers used for our experiments on the ImageNet-1k dataset are a tiny (ViT-T), small (ViT-S), base (ViT-B), and large (ViT-L). The dimensionality of the representations produced by these models are given in \hyperref[tab:repr-dims]{Table~\ref{tab:repr-dims}}.

\begin{table}[h!]
\centering
\caption{Vision transformer sizes evaluated in this work.}
\label{tab:repr-dims}
\begin{tabular}{|c|c|}
\hline
 Model & Repr.~Dimensionality \\ 
 \hline
 ViT-T & 192 \\ \hline
 ViT-S & 384 \\ \hline
 ViT-B & 768 \\ \hline
 ViT-L & 1024 \\ \hline
\end{tabular}
\end{table}

\begin{figure}[h]
    \centering
    \includegraphics[width=\columnwidth]{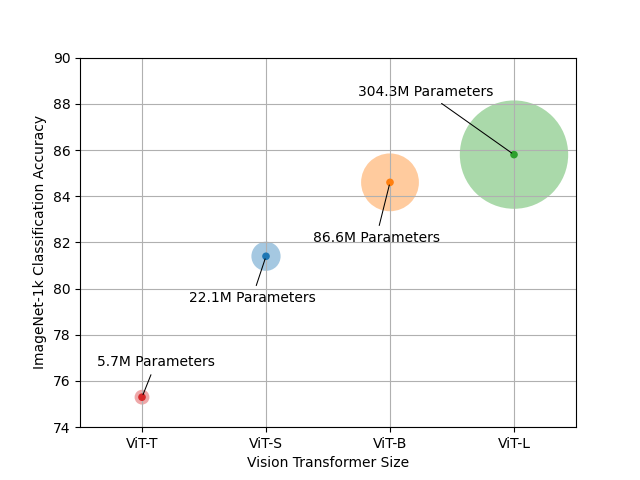}
    \caption{ImageNet-1k classification accuracy of pre-trained vision transformers and classification heads using checkpoints obtained from \cite{rw2019timm}. The shaded area around each data point represents the number of parameters in the vision transformer.}
    \label{fig:parameter-validation}
\end{figure}

We validate the performance of the model checkpoints on the ImageNet-1k dataset using the original classification head for each model. For use in our experiments, we remove this classification head from each vision transformer and evaluate the representations produced by these models (\hyperref[def:img-representation]{Definition~\ref{def:img-representation}}) using a linear classifier (\hyperref[def:linear-classifier]{Definition~\ref{def:linear-classifier}}). \hyperref[fig:parameter-validation]{Figure~\ref{fig:parameter-validation}} shows the ImageNet-1k classification accuracy of our pre-trained vision transformers using their original classification heads. 

To show the performance of our method on datasets other than ImageNet-1k, we use vision transformers trained using the CLIP method described in \cite{radford2021learningtransferablevisualmodels}. Vision transformers trained in this manner learn representations that generalize well to unseen data without dataset-specific fine-tuning. We obtain checkpoints of model parameters for these vision transformers from \cite{ilharco_gabriel_2021_5143773}. These models are pre-trained on the Laion-400M dataset \cite{schuhmann2021laion400mopendatasetclipfiltered}. We evaluate both a ViT-B and ViT-L trained using the CLIP method which produce representations with dimensionalities of 512 and 768, respectively.\footnote{The dimensionalities of the representations produced by these models differ from the ViT-B and ViT-L used for ImageNet-1k experiments due to the addition of projection heads used in CLIP style training. See \cite{radford2021learningtransferablevisualmodels}.} 
See \hyperref[sec:app-model-details]{Appendix ~\ref{sec:app-model-details}} for more details about the pre-trained vision transformers used in our experiments.

\subsection{Datasets}

We show the effectiveness of our method on representations of the ImageNet-1k \cite{deng2009imagenet}, CIFAR-100 \cite{Krizhevsky2009LearningML}, Caltech-101 \cite{li_andreeto_ranzato_perona_2022}, MNIST \cite{6296535}, and Oxford-IIIT Pets \cite{parkhi12a} datasets. For the ImageNet-1k, CIFAR-100, and MNIST datasets we use 10\% of the training samples, with an equal number of images per class, in our experiments. Due to their much smaller size, we use the full Caltech-101 and Oxford-IIIT Pets datasets. For all datasets, we report the classification accuracy on the full validation set. 

\begin{figure}[h]
    \centering
    \includegraphics[width=\columnwidth]{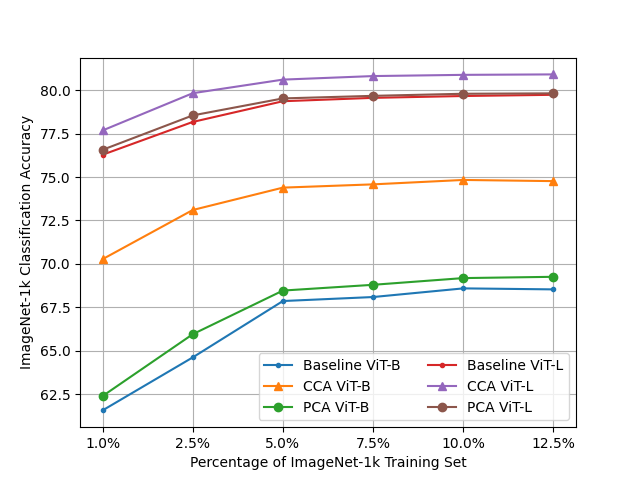}
    \caption{The ImageNet-1k classification accuracy of a linear probe trained using representations of varying amounts of the ImageNet-1k training data produced by a ViT-B and ViT-L.}
    \label{fig:train_set_size}
\end{figure}

\hyperref[fig:train_set_size]{Figure~\ref{fig:train_set_size}} shows the ImageNet-1k accuracy of a linear probe trained on representations of increasing amounts of the ImageNet-1k training set. These results show that using more than 10\% of the ImageNet-1k training samples leads to diminishing returns in accuracy and adds additional computational cost to showing the effectiveness of our method. 

\subsection{CCA Subspace Projection}
Given pre-trained vision transformers $M_X$ and $M_Y$ and a dataset $I$ containing $N$ images, we construct matrices $I_X$ and $I_Y$ containing representations of images in $I$ produced by $M_X$ and $M_Y$. Using these matrices, we find two linear transformations $U, V = \text{CCA}(I_X, I_Y)$. Matrices $U$ and $V$ define the linear transformations used to project centered and whitened versions of $I_X$ and $I_Y$ into a subspace in which their dimensions are correlated. Formal definitions are as follows. 

\definition{
    \label{def:representation-mats}
    Let $I_X = M_X(I)$ and $I_Y = M_Y(I)$ denote matrices whose columns are representations of the $N$ images in $I$ produced by models $M_X$ and $M_Y$, respectively. Thus, $I_X \in \mathbb{R}^{d_X \times N}$ and $I_Y \in \mathbb{R}^{d_Y \times N}$ where $d_X$ and $d_Y$ are the dimensionalities of the representations produced by $M_X$ and $M_Y$. 
}

\definition{
    \label{def:cca-transforms}
    Let $U \in \mathbb{R}^{d \times d_X}$ and $V \in \mathbb{R}^{d \times d_Y}$ define linear transformations that project representations $I_X$ and $I_Y$ into a new $d=\min(d_X, d_Y)$ dimensional space. Then
    \begin{gather*}
        U, \Lambda, V^T = \text{CCA}(I_X, I_Y)
    \end{gather*}
    where $\Lambda \in \mathbb{R}^{d \times d}$ denotes a diagonal matrix whose elements $1 \geq \lambda_1 \geq ... \geq \lambda_d \geq 0$ are the correlation coefficients of the dimensions of the projected representations.
}

\definition{
\label{def:cca-projection}
    Let matrices $I_X', I_Y' \in \mathbb{R}^{d \times N}$ denote the representation matrices after projection. Then
    \begin{align*}
        I_X' = U\, \Sigma_{X}^{-1/2} (I_X - \mu_{X}) \\
        I_Y' = V\, \Sigma_{Y}^{-1/2} (I_Y - \mu_{Y})
    \end{align*}
    where $\Sigma_X$ and $\Sigma_Y$ are covariance matrices for $I_X$ and $I_Y$ and $\mu_X$ and $\mu_Y$ are mean vectors. See Appendix A of \cite{Krizhevsky2009LearningML} for a detailed explanation of the ZCA whitening transformation. 
}

\begin{table*}[t!]
\centering
\caption{Reducing dimensionality of larger ViT representations using smaller ViT representations}
\label{tab:reduce-dim}
\begin{tabular}{|c|c|c|c|c|c|c|c|c|}
\hline
Model & CCA Partner & Baseline & PCA & CCA & Orig. Dim. & Proj. Dim. & Dim. $\Delta$ \\ \hline

ViT-S & ViT-T & 63.3\% & 61.9\% & \textbf{65.9\%} & 384 & 192 & -50\% \\ \hline

\multirow{2}{*}{ViT-B} & ViT-T & \multirow{2}{*}{68.6\%} & 65.8\% & \textbf{69.9\%} & 768 & 192 & -75\% \\ \cline{2-2} \cline{4-8} 
 & ViT-S & & 68.4\% & \textbf{73.1\%} & 768 & 384 & -50\% \\ \hline
 
\multirow{3}{*}{ViT-L} & ViT-T & \multirow{2}{*}{\textbf{79.6\%}} & 78.1\% & 76.3\% & 1024 & 192 & -81.3\% \\ \cline{2-2} \cline{4-8} 
 & ViT-S & & 79.4\% & 79.3\% & 1024 & 384 & -62.5\% \\ \cline{2-8} 
 & ViT-B & 79.6\% & 79.8\% & \textbf{80.9\%} & 1024 & 768 & -25\% \\ \hline
\end{tabular}
\end{table*}

\subsection{Representation Evaluation}

\textbf{Representation selection and distillation:} To evaluate our method on ImageNet-1k, we run experiments that compute CCA and apply the resulting transformations to representations produced by each pair of the four vision transformers listed in \hyperref[tab:repr-dims]{Table~\ref{tab:repr-dims}}. For a given pair of vision transformers $M_X, M_Y$, representations are left unchanged for a baseline evaluation or projected using linear transformations found via either PCA or CCA. If the representations are projected using PCA, the resulting representations have the same number of dimensions as they would if they were projected using CCA; i.e. to the smaller dimensionality. Note that compute of CCA and PCA transformation matrices is on training samples.

After projection, a linear classifier (\hyperref[def:linear-classifier]{Definition~\ref{def:linear-classifier}}) is trained on $I'_X$ and $I'_Y$; for baseline experiments, $I_X$ and $I_Y$ are used. This classifier is trained for 100 epochs using stochastic gradient descent optimization with a learning rate of 0.01, a momentum of 0.9, and a weight decay of 5e-4. We use the accuracy on the ImageNet-1k validation set to measure the quality of representations.

\textbf{Technique generalization:} For experiments on datasets other than ImageNet-1k, the same procedure is followed, except we only conduct experiments using a ViT-B and ViT-L that were trained using the CLIP objective. When using the Caltech-101 dataset, the linear classifier is trained for 300 epochs instead of the 100 used elsewhere as we notice that convergence takes longer for this particular dataset. For the ImageNet-1k, CIFAR-100, and MNIST datasets we use a batch size of 128 when training the linear classifier, and we use a batch size of 64 when using the Caltech-101 and Oxford-IIIT Pets datasets due to the much smaller number of samples in these two datasets. 

\textbf{Structure transfer:} To further show the effectiveness of our projection method, we conduct an experiment in which we consider a pair of vision transformers of the same size. Both models are pre-trained on ImageNet-21k but only one is fine-tuned on ImageNet-1k. We apply the previously described experimental procedure to representations produced by these models. The goal of this experiment is to show that our projection method can improve the ImageNet-1k classification accuracy of the representations produced by the model that was not fine-tuned on ImageNet-1k.

In general, our experimental procedure is as follows: 
\begin{enumerate}
    \item Choose two, pre-trained, vision transformers that take an image as input and produce a high-dimensional representation of that image. 
        
    \item For each of the models, create two representation matrices. One which contains representations of the training data and another which contains representations of the validation data.

    \item Project the representations using PCA or CCA, or keep them unchanged for baseline evaluation. 

    \item For each model, train a linear classifier using the representations of the training data. The performance of the linear classifier is evaluated using the representations of the validation data.
\end{enumerate}

For each experiment, we repeat the above process 5 times using different seeds for dataset splitting and linear classifier initialization. We report the mean accuracy of these 5 trials. The associated standard deviation is $\leq 0.01\%$ for all experiments and therefore not reported in the following tables.

\section{Results}
\label{sec:results}

We evaluate CCA-based projections across three complementary regimes: (i) representation selection via dimensionality reduction, (ii) fixed-dimensional representation refinement, and (iii) post-hoc transfer of task-relevant structure without additional learning. Table~\ref{tab:reduce-dim} shows that CCA consistently outperforms both unprojected representations and PCA-based dimensionality reduction. Averaged across model pairs, CCA improves classification accuracy by 1.02\% (up to 4.5\%) over the baseline and by 2.0\% (up to 4.7\%) over PCA, while achieving an average dimensionality reduction of 57\%.

Table~\ref{tab:same-dim} further demonstrates that CCA can refine fixed-dimensional embeddings by distilling structure from higher-capacity models into a shared projection space, without modifying or retraining either model. Across all evaluated model combinations, CCA improves classification accuracy with an average gain of approximately 8\% (up to 10.5\%).

\begin{table}[h]
\centering
\caption{Knowledge distillation from larger ViT to smaller ViT representations. Note that dimensionality is unchanged.}
\label{tab:same-dim}
\begin{tabular}{|c|c|c|c|c|}
\hline
Model & CCA Partner & Dim & Baseline & CCA \\ \hline

\multirow{3}{*}{ViT-T} & ViT-S &  \multirow{3}{*}{192} & \multirow{3}{*}{34.2\%} & 44.7\% \\ \cline{2-2} \cline{5-5} 
 & ViT-B & & & \textbf{44.8\%} \\ \cline{2-2} \cline{5-5} 
 & ViT-L & & & \textbf{44.8\%} \\ \hline
 
\multirow{2}{*}{ViT-S} & ViT-B & \multirow{2}{*}{384} & \multirow{2}{*}{63.3\%} & \textbf{68.4\%} \\ \cline{2-2} \cline{5-5} 
 & ViT-L & &  & \textbf{68.4\%} \\ \hline
 
ViT-B & ViT-L & 768 & 68.6\%  & \textbf{74.7\%} \\ \hline
\end{tabular}
\end{table}

Table~\ref{tab:finetune-transfer} evaluates CCA as a post-hoc structure transfer operator. Across all Vision Transformer (ViT) model combinations and scales, CCA improves classification performance for ImageNet-21k–pretrained representations, regardless of whether the evaluated model is fine-tuned on ImageNet-1k. When the CCA partner is fine-tuned on the downstream task, representations from non–fine-tuned models improve by an average of 8.2\% (up to 12.6\%), while fine-tuned models themselves improve by an average of 5.9\% (up to 10.8\%).

\begin{table}[]
\centering
\caption{Structure transfer from fine-tuned ViT to non-finetuned ViT representations. Note that dimensionality is unchanged.}
\label{tab:finetune-transfer}
\begin{tabular}{|c|c|c|c|c|}
\hline
Model & Dim & Finetuned & Baseline & CCA \\ \hline
\multirow{2}{*}{ViT-T} & \multirow{2}{*}{192} & No & 29.0\% & 41.6\% \\
 & & Yes & 34.2\% & 45.0\% \\ \hline
\multirow{2}{*}{ViT-S} &\multirow{2}{*}{384} & No & 58.8\% & 66.2\% \\
 & & Yes & 63.3\% & 68.4\% \\ \hline
\multirow{2}{*}{ViT-B} & \multirow{2}{*}{768} & No & 62.0\% & 71.9\% \\
 & & Yes & 68.6\% & 74.7\% \\ \hline
\multirow{2}{*}{ViT-L} & \multirow{2}{*}{1024} & No & 76.3\% & 79.3\% \\
& & Yes & 79.6\% & 81.2\% \\ \hline
\end{tabular}
\end{table}

\hyperref[tab:multidataset-results]{Table~\ref{tab:multidataset-results}} reports the performance of CCA-based projections across multiple datasets beyond ImageNet-1k. On CIFAR-100 and MNIST, CCA consistently improves classification accuracy by an average of 4.15\%, substantially outperforming PCA-based projections, which yield an average gain of 1.2\%. On the Oxford-IIIT Pets dataset, CCA provides a smaller but consistent improvement of 0.35\%, while PCA-based projections yield larger gains on this dataset, improving accuracy by an average of 1.27\%.

\begin{table*}[h!]
\centering
\caption{Linear probing accuracy of representations of various image classification datasets when using vision transformer models trained as CLIP image encoders on LAION-400M.}
\label{tab:multidataset-results}
\begin{tabular}{|c|c|c|ccccc|}
\hline
\multirow{2}{*}{Model} & \multirow{2}{*}{Method} & \multicolumn{1}{l|}{\multirow{2}{*}{Dim.}} & \multicolumn{5}{c|}{Dataset} \\ \cline{4-8} 
 &  & \multicolumn{1}{l|}{} & \multicolumn{1}{c|}{IN-1k} & \multicolumn{1}{c|}{CIFAR-100} & \multicolumn{1}{c|}{MNIST} & \multicolumn{1}{c|}{Caltech-101} & Oxford-IIIT Pets \\ \hline
\multirow{3}{*}{ViT-B} & Baseline & 512 & \multicolumn{1}{c|}{58.89\%} & \multicolumn{1}{c|}{69.72\%} & \multicolumn{1}{c|}{94.8\%} & \multicolumn{1}{c|}{86.29\%} & 86.4\% \\ \cline{2-8} 
 
 & PCA & 512 & \multicolumn{1}{c|}{59.46\%} & \multicolumn{1}{c|}{70.91\%} & \multicolumn{1}{c|}{96.4\%} & \multicolumn{1}{c|}{\textbf{91.59\%}} & \textbf{87.91\%} \\ \cline{2-8} 

 & CCA & 512 & \multicolumn{1}{c|}{\textbf{62.38\%}} & \multicolumn{1}{c|}{\textbf{75.8\%}} & \multicolumn{1}{c|}{\textbf{98.09\%}} & \multicolumn{1}{c|}{55.22\%} & 86.81\% \\ \hline
 
\multirow{3}{*}{ViT-L} & Baseline & 768 & \multicolumn{1}{c|}{65.9\%} & \multicolumn{1}{c|}{76.59\%} & \multicolumn{1}{c|}{95.58\%} & \multicolumn{1}{c|}{89.33\%} & 90.4\% \\ \cline{2-8} 
 
 & PCA & 512 & \multicolumn{1}{c|}{66.32\%} & \multicolumn{1}{c|}{77.28\%} & \multicolumn{1}{c|}{96.88\%} & \multicolumn{1}{c|}{\textbf{92.73\%}} & \textbf{91.43\%} \\ \cline{2-8} 
 
 & CCA & 512 & \multicolumn{1}{c|}{\textbf{68.11\%}} & \multicolumn{1}{c|}{\textbf{81.14\%}} & \multicolumn{1}{c|}{\textbf{98.27\%}} & \multicolumn{1}{c|}{58.54\%} & 90.68\% \\ \hline
\end{tabular}
\end{table*}

On the Caltech-101 dataset, CCA-based projections reduce downstream classification accuracy for both models, whereas PCA-based projections lead to improvements. Notably, Caltech-101 is the only dataset in our evaluation with substantial class imbalance. This behavior reveals a structured failure mode of CCA-based alignment: when class imbalance is severe, covariance estimates used for whitening may be dominated by majority classes, limiting their ability to capture shared semantic structure. Rather than a limitation of the method’s applicability, this result highlights the sensitivity of cross-model agreement to data distribution and suggests that CCA can serve as a diagnostic tool for identifying settings in which representation alignment is ill-posed.

\begin{figure}[h]
    \centering
    \includegraphics[width=\columnwidth]{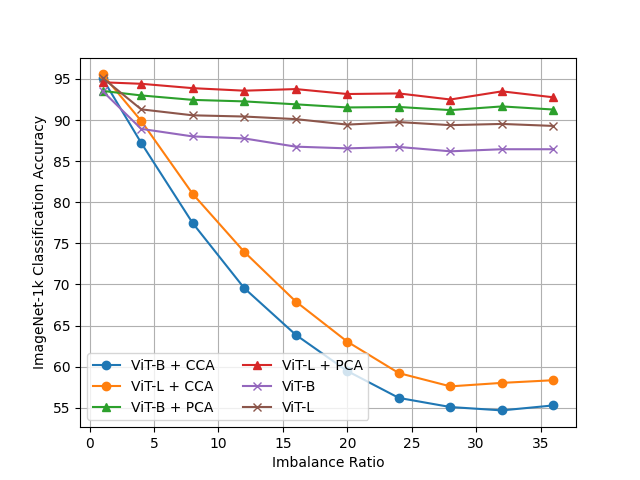}
    \caption{Classification accuracy of a linear probe trained on representations of increasingly imbalanced subsets of the Caltech-101 dataset. These representations are produced by a ViT-B and ViT-L trained using the CLIP objective. The X-axis is the maximum allowed ratio between the number of samples in the least common class and the number of samples in the most common class.}
    \label{fig:imbalance_v_acc}
\end{figure}

To explicitly examine the sensitivity of CCA-based projections to class imbalance, we repeat our experimental procedure on progressively imbalanced subsets of the Caltech-101 training data. In all experiments, the validation set retains the original dataset imbalance; only the imbalance ratio of the training set is varied. As shown in \hyperref[fig:imbalance_v_acc]{Figure~\ref{fig:imbalance_v_acc}}, increasing class imbalance in the training data leads to a monotonic decline in classification accuracy for CCA-projected representations. This trend confirms that CCA-based alignment is sensitive to distributional skew in the data used to estimate cross-model covariance. The imbalance ratio in \hyperref[fig:imbalance_v_acc]{Figure~\ref{fig:imbalance_v_acc}} is defined as $R = \frac{N_1}{N_2}$, where $N_1$ and $N_2$ denote the number of samples in the most frequent and least frequent classes, respectively.

\section{Discussion}

Our results show that canonical correlation analysis provides a simple yet principled mechanism for post-hoc representation selection based on cross-model agreement. Rather than prioritizing directions of high variance within a single embedding space, CCA isolates representation components that are mutually predictable across independently trained models, which we find to be a reliable proxy for task-relevant semantic structure. This perspective explains why CCA-based projections can simultaneously reduce dimensionality and improve downstream performance across a range of settings (Tables \ref{tab:reduce-dim}–\ref{tab:same-dim}). Importantly, this signal arises from agreement across models rather than statistics of any single representation.

The gains observed under fixed-dimensional projections further indicate that CCA acts as a representation refinement operator rather than a compression mechanism alone. As shown in Table \ref{tab:same-dim}, aligning representations using agreement filters weakly shared directions while preserving structure useful across models, leading to consistent accuracy improvements without retraining or supervision. Importantly, this refinement is symmetric: both aligned representations benefit from projection. This behavior distinguishes CCA-based refinement from teacher–student distillation, which relies on asymmetric optimization and explicit supervision, and instead suggests that cross-model agreement saturates with capacity, and that effective alignment does not require extreme capacity mismatches.

Results from the fine-tuned / non–fine-tuned setting (Table \ref{tab:finetune-transfer}) further show that cross-model alignment enables post-hoc transfer of task-specific structure. When one model is fine-tuned on a downstream task, CCA-based projections allow a frozen partner model to benefit from this adaptation without access to labels or gradient-based updates. This suggests that fine-tuning reshapes representation geometry in ways that remain partially predictable from related pretrained models, enabling indirect knowledge transfer through alignment alone. In practice, this supports the reuse of frozen or deployment-constrained models and highlights CCA as a lightweight mechanism for propagating downstream improvements without retraining.

While CCA-based alignment exhibits effects related to distillation, representation selection, and knowledge transfer, it operates under a fundamentally different set of assumptions than methods designed specifically for these tasks. Most state-of-the-art approaches in these areas rely on retraining, task-specific losses, or asymmetric teacher–student setups, whereas our method is training-free, symmetric, and applied post-hoc to frozen representations. As a result, direct quantitative comparisons are not meaningful, and our results should be interpreted as evidence of a distinct, complementary operating regime rather than a replacement for specialized training-based techniques.

\begin{figure}[h]
    \centering
    \includegraphics[width=\columnwidth]{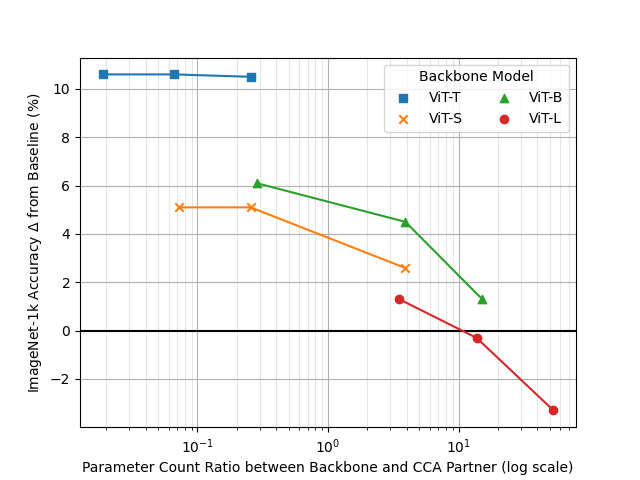}
    \caption{Comparing CCA improvements over baseline as a function of the parameter ratio between backbone and partner models.}
    \label{fig:capacity}
\end{figure}

To better understand how cross-model agreement relates to model capacity, we analyze how the effectiveness of CCA-based projections varies with the relative capacities of the backbone and partner models. As summarized in Figure \ref{fig:capacity}, the results reveal a clear diminishing-returns behavior: for smaller backbone models (e.g., ViT-T and ViT-S), increasing partner capacity beyond a certain point yields little additional improvement, while for larger backbones (e.g., ViT-B and ViT-L), performance degrades approximately logarithmically as partner capacity is reduced. Together with the results in Tables \ref{tab:reduce-dim} and \ref{tab:same-dim}, this suggests that CCA-based alignment is most effective when model capacities differ by a moderate factor, large enough to expose complementary structure, but close enough for shared abstractions to remain reliably identifiable through cross-covariance estimation.

Performance degradation under class imbalance provides additional insight into when agreement-based alignment is ill-posed. On Caltech-101, and under increasing imbalance ratios (Figure \ref{fig:imbalance_v_acc}, Table \ref{tab:multidataset-results}), CCA-based projections degrade while PCA-based projections remain effective. Because CCA relies on reliable covariance and cross-covariance estimates, severe imbalance causes these estimates to be dominated by majority classes, obscuring shared structure across models. Rather than a limitation of the approach, this behavior highlights a diagnostic property of agreement-based methods, indicating when representation alignment is unreliable due to data distribution.

\section{Conclusion}
\label{sec:conclusion}

This work demonstrates that canonical correlation analysis provides a simple and principled mechanism for post-hoc representation selection based on cross-model agreement. By isolating representation components that are mutually predictable across independently trained models, CCA enables both effective dimensionality reduction and fixed-dimensional refinement, often improving downstream performance without retraining or architectural modification.

Unlike variance-based methods such as PCA, which rely on statistics from a single embedding space, CCA leverages agreement between multiple representations to guide selection. This shifts representation selection from a data-dependent heuristic to a model-driven operation that exploits structure inherent to learned representations. Our results show that this perspective supports compression, refinement, and post-hoc transfer of task-specific structure, while also revealing structured failure modes under severe class imbalance. Future work could explore extensions of this framework, including robust covariance estimation, class-balanced alignment, and interactions between agreement-based post-hoc alignment and training-based distillation or compression methods; however, the focus of this work is to characterize what can be achieved without retraining, using cross-model agreement alone.

As modern vision pipelines increasingly rely on collections of pretrained models trained under different objectives and constraints, lightweight mechanisms for aligning and reusing representations become essential. Because CCA-based alignment is linear, model-agnostic, and training-free, it provides a practical tool for representation interoperability across models and tasks. While our experiments focus on Vision Transformers, the underlying principle of agreement-based representation selection is broadly applicable and suggests a general strategy for managing and reusing learned representations in large-scale vision systems.

\bibliographystyle{IEEEtran}
\bibliography{ref}

@misc{dosovitskiy2021imageworth16x16words,
      title={An Image is Worth 16x16 Words: Transformers for Image Recognition at Scale}, 
      author={Alexey Dosovitskiy and Lucas Beyer and Alexander Kolesnikov and Dirk Weissenborn and Xiaohua Zhai and Thomas Unterthiner and Mostafa Dehghani and Matthias Minderer and Georg Heigold and Sylvain Gelly and Jakob Uszkoreit and Neil Houlsby},
      year={2021},
      eprint={2010.11929},
      archivePrefix={arXiv},
      primaryClass={cs.CV},
      url={https://arxiv.org/abs/2010.11929}, 
}

@article{da6385d2-9c65-3860-bbcd-b821fdff69ff,
 ISSN = {00063444},
 URL = {http://www.jstor.org/stable/2333955},
 author = {Harold Hotelling},
 journal = {Biometrika},
 number = {3/4},
 pages = {321--377},
 publisher = {[Oxford University Press, Biometrika Trust]},
 title = {Relations Between Two Sets of Variates},
 urldate = {2025-11-12},
 volume = {28},
 year = {1936}
}

@misc{raghu2017svccasingularvectorcanonical,
      title={SVCCA: Singular Vector Canonical Correlation Analysis for Deep Learning Dynamics and Interpretability}, 
      author={Maithra Raghu and Justin Gilmer and Jason Yosinski and Jascha Sohl-Dickstein},
      year={2017},
      eprint={1706.05806},
      archivePrefix={arXiv},
      primaryClass={stat.ML},
      url={https://arxiv.org/abs/1706.05806}, 
}

@misc{morcos2018insightsrepresentationalsimilarityneural,
      title={Insights on representational similarity in neural networks with canonical correlation}, 
      author={Ari S. Morcos and Maithra Raghu and Samy Bengio},
      year={2018},
      eprint={1806.05759},
      archivePrefix={arXiv},
      primaryClass={stat.ML},
      url={https://arxiv.org/abs/1806.05759}, 
}

@misc{radford2021learningtransferablevisualmodels,
      title={Learning Transferable Visual Models From Natural Language Supervision}, 
      author={Alec Radford and Jong Wook Kim and Chris Hallacy and Aditya Ramesh and Gabriel Goh and Sandhini Agarwal and Girish Sastry and Amanda Askell and Pamela Mishkin and Jack Clark and Gretchen Krueger and Ilya Sutskever},
      year={2021},
      eprint={2103.00020},
      archivePrefix={arXiv},
      primaryClass={cs.CV},
      url={https://arxiv.org/abs/2103.00020}, 
}

@misc{paszke2019pytorchimperativestylehighperformance,
      title={PyTorch: An Imperative Style, High-Performance Deep Learning Library}, 
      author={Adam Paszke and Sam Gross and Francisco Massa and Adam Lerer and James Bradbury and Gregory Chanan and Trevor Killeen and Zeming Lin and Natalia Gimelshein and Luca Antiga and Alban Desmaison and Andreas Köpf and Edward Yang and Zach DeVito and Martin Raison and Alykhan Tejani and Sasank Chilamkurthy and Benoit Steiner and Lu Fang and Junjie Bai and Soumith Chintala},
      year={2019},
      eprint={1912.01703},
      archivePrefix={arXiv},
      primaryClass={cs.LG},
      url={https://arxiv.org/abs/1912.01703}, 
}

@inproceedings{deng2009imagenet,
  title={Imagenet: A large-scale hierarchical image database},
  author={Deng, Jia and Dong, Wei and Socher, Richard and Li, Li-Jia and Li, Kai and Fei-Fei, Li},
  booktitle={2009 IEEE conference on computer vision and pattern recognition},
  pages={248--255},
  year={2009},
  organization={IEEE}
}

@misc{rw2019timm,
  author = {Ross Wightman},
  title = {PyTorch Image Models},
  year = {2019},
  publisher = {GitHub},
  journal = {GitHub repository},
  doi = {10.5281/zenodo.4414861},
  howpublished = {\url{https://github.com/rwightman/pytorch-image-models}}
}

@misc{assran2023selfsupervisedlearningimagesjointembedding,
      title={Self-Supervised Learning from Images with a Joint-Embedding Predictive Architecture}, 
      author={Mahmoud Assran and Quentin Duval and Ishan Misra and Piotr Bojanowski and Pascal Vincent and Michael Rabbat and Yann LeCun and Nicolas Ballas},
      year={2023},
      eprint={2301.08243},
      archivePrefix={arXiv},
      primaryClass={cs.CV},
      url={https://arxiv.org/abs/2301.08243}, 
}

@inproceedings{chaudhuri2009multi,
  title={Multi-view clustering via canonical correlation analysis},
  author={Chaudhuri, Kamalika and Kakade, Sham M and Livescu, Karen and Sridharan, Karthik},
  booktitle={Proceedings of the 26th annual international conference on machine learning},
  pages={129--136},
  year={2009}
}

@misc{wang2016deepmultiviewrepresentationlearning,
      title={On Deep Multi-View Representation Learning: Objectives and Optimization}, 
      author={Weiran Wang and Raman Arora and Karen Livescu and Jeff Bilmes},
      year={2016},
      eprint={1602.01024},
      archivePrefix={arXiv},
      primaryClass={cs.LG},
      url={https://arxiv.org/abs/1602.01024}, 
}

@InProceedings{pmlr-v28-andrew13,
  title = 	 {Deep Canonical Correlation Analysis},
  author = 	 {Andrew, Galen and Arora, Raman and Bilmes, Jeff and Livescu, Karen},
  booktitle = 	 {Proceedings of the 30th International Conference on Machine Learning},
  pages = 	 {1247--1255},
  year = 	 {2013},
  editor = 	 {Dasgupta, Sanjoy and McAllester, David},
  volume = 	 {28},
  number =       {3},
  series = 	 {Proceedings of Machine Learning Research},
  address = 	 {Atlanta, Georgia, USA},
  month = 	 {17--19 Jun},
  publisher =    {PMLR},
  url = 	 {https://proceedings.mlr.press/v28/andrew13.html}
}

@misc{benton2017deepgeneralizedcanonicalcorrelation,
      title={Deep Generalized Canonical Correlation Analysis}, 
      author={Adrian Benton and Huda Khayrallah and Biman Gujral and Dee Ann Reisinger and Sheng Zhang and Raman Arora},
      year={2017},
      eprint={1702.02519},
      archivePrefix={arXiv},
      primaryClass={cs.LG},
      url={https://arxiv.org/abs/1702.02519}, 
}

@ARTICLE{9428614,
  author={Wong, Hok Shing and Wang, Li and Chan, Raymond and Zeng, Tieyong},
  journal={IEEE Transactions on Big Data}, 
  title={Deep Tensor CCA for Multi-View Learning}, 
  year={2022},
  volume={8},
  number={6},
  pages={1664-1677},
  doi={10.1109/TBDATA.2021.3079234}}

@misc{chen2020simpleframeworkcontrastivelearning,
      title={A Simple Framework for Contrastive Learning of Visual Representations}, 
      author={Ting Chen and Simon Kornblith and Mohammad Norouzi and Geoffrey Hinton},
      year={2020},
      eprint={2002.05709},
      archivePrefix={arXiv},
      primaryClass={cs.LG},
      url={https://arxiv.org/abs/2002.05709}, 
}

@misc{he2020momentumcontrastunsupervisedvisual,
      title={Momentum Contrast for Unsupervised Visual Representation Learning}, 
      author={Kaiming He and Haoqi Fan and Yuxin Wu and Saining Xie and Ross Girshick},
      year={2020},
      eprint={1911.05722},
      archivePrefix={arXiv},
      primaryClass={cs.CV},
      url={https://arxiv.org/abs/1911.05722}, 
}

@misc{grill2020bootstraplatentnewapproach,
      title={Bootstrap your own latent: A new approach to self-supervised Learning}, 
      author={Jean-Bastien Grill and Florian Strub and Florent Altché and Corentin Tallec and Pierre H. Richemond and Elena Buchatskaya and Carl Doersch and Bernardo Avila Pires and Zhaohan Daniel Guo and Mohammad Gheshlaghi Azar and Bilal Piot and Koray Kavukcuoglu and Rémi Munos and Michal Valko},
      year={2020},
      eprint={2006.07733},
      archivePrefix={arXiv},
      primaryClass={cs.LG},
      url={https://arxiv.org/abs/2006.07733}, 
}

@misc{chen2020exploringsimplesiameserepresentation,
      title={Exploring Simple Siamese Representation Learning}, 
      author={Xinlei Chen and Kaiming He},
      year={2020},
      eprint={2011.10566},
      archivePrefix={arXiv},
      primaryClass={cs.CV},
      url={https://arxiv.org/abs/2011.10566}, 
}

@article{10.1162/153244303768966085,
author = {Bach, Francis R. and Jordan, Michael I.},
title = {Kernel independent component analysis},
year = {2003},
issue_date = {3/1/2003},
publisher = {JMLR.org},
volume = {3},
number = {null},
issn = {1532-4435},
url = {https://doi.org/10.1162/153244303768966085},
doi = {10.1162/153244303768966085},
journal = {J. Mach. Learn. Res.},
month = mar,
pages = {1–48},
numpages = {48}
}

@article{10.5555/1248659.1248673,
author = {Fukumizu, Kenji and Bach, Francis R. and Gretton, Arthur},
title = {Statistical Consistency of Kernel Canonical Correlation Analysis},
year = {2007},
issue_date = {5/1/2007},
publisher = {JMLR.org},
volume = {8},
issn = {1532-4435},
journal = {J. Mach. Learn. Res.},
month = may,
pages = {361–383},
numpages = {23}
}

@ARTICLE{Nielsen2002-hu,
  title     = "Multiset canonical correlations analysis and multispectral,
               truly multitemporal remote sensing data",
  author    = "Nielsen, Allan Aasbjerg",
  journal   = "IEEE Trans. Image Process.",
  publisher = "Institute of Electrical and Electronics Engineers (IEEE)",
  volume    =  11,
  number    =  3,
  pages     = "293--305",
  year      =  2002,
  copyright = "https://ieeexplore.ieee.org/Xplorehelp/downloads/license-information/IEEE.html",
  language  = "en"
}

@misc{Krizhevsky2009LearningML,
  title={Learning Multiple Layers of Features from Tiny Images},
  author={Alex Krizhevsky},
  year={2009},
  url={https://api.semanticscholar.org/CorpusID:18268744}
}

@misc{steiner2022trainvitdataaugmentation,
      title={How to train your ViT? Data, Augmentation, and Regularization in Vision Transformers}, 
      author={Andreas Steiner and Alexander Kolesnikov and Xiaohua Zhai and Ross Wightman and Jakob Uszkoreit and Lucas Beyer},
      year={2022},
      eprint={2106.10270},
      archivePrefix={arXiv},
      primaryClass={cs.CV},
      url={https://arxiv.org/abs/2106.10270}, 
}

@misc{he2021maskedautoencodersscalablevision,
      title={Masked Autoencoders Are Scalable Vision Learners}, 
      author={Kaiming He and Xinlei Chen and Saining Xie and Yanghao Li and Piotr Dollár and Ross Girshick},
      year={2021},
      eprint={2111.06377},
      archivePrefix={arXiv},
      primaryClass={cs.CV},
      url={https://arxiv.org/abs/2111.06377}, 
}

@misc{caron2021emergingpropertiesselfsupervisedvision,
      title={Emerging Properties in Self-Supervised Vision Transformers}, 
      author={Mathilde Caron and Hugo Touvron and Ishan Misra and Hervé Jégou and Julien Mairal and Piotr Bojanowski and Armand Joulin},
      year={2021},
      eprint={2104.14294},
      archivePrefix={arXiv},
      primaryClass={cs.CV},
      url={https://arxiv.org/abs/2104.14294}, 
}

@article{Hotelling_1933, title={Analysis of a complex of statistical variables into principal components.}, volume={24}, DOI={10.1037/h0071325}, number={6}, journal={Journal of Educational Psychology}, author={Hotelling, Harold}, year={1933}, month={Sep}, pages={417–441}}

@article{fisher1936lda,
  author  = {Fisher, R. A.},
  title   = {The Use of Multiple Measurements in Taxonomic Problems},
  journal = {Annals of Eugenics},
  volume  = {7},
  number  = {2},
  pages   = {179--188},
  year    = {1936}
}

@InProceedings{10.1007/BFb0020217,
author="Sch{\"o}lkopf, Bernhard
and Smola, Alexander
and M{\"u}ller, Klaus-Robert",
editor="Gerstner, Wulfram
and Germond, Alain
and Hasler, Martin
and Nicoud, Jean-Daniel",
title="Kernel principal component analysis",
booktitle="Artificial Neural Networks --- ICANN'97",
year="1997",
publisher="Springer Berlin Heidelberg",
address="Berlin, Heidelberg",
pages="583--588",
isbn="978-3-540-69620-9"
}

@inproceedings{NIPS1999_c0d0e461,
 author = {Roth, Volker and Steinhage, Volker},
 booktitle = {Advances in Neural Information Processing Systems},
 editor = {S. Solla and T. Leen and K. M\"{u}ller},
 pages = {},
 publisher = {MIT Press},
 title = {Nonlinear Discriminant Analysis Using Kernel Functions},
 url = {https://proceedings.neurips.cc/paper_files/paper/1999/file/c0d0e461de8d0024aebcb0a7c68836df-Paper.pdf},
 volume = {12},
 year = {1999}
}

@misc{ilharco_gabriel_2021_5143773,
  author       = {Ilharco, Gabriel and
                  Wortsman, Mitchell and
                  Wightman, Ross and
                  Gordon, Cade and
                  Carlini, Nicholas and
                  Taori, Rohan and
                  Dave, Achal and
                  Shankar, Vaishaal and
                  Namkoong, Hongseok and
                  Miller, John and
                  Hajishirzi, Hannaneh and
                  Farhadi, Ali and
                  Schmidt, Ludwig},
  title        = {OpenCLIP},
  month        = jul,
  year         = 2021,
  note         = {If you use this software, please cite it as below.},
  publisher    = {Zenodo},
  version      = {0.1},
  doi          = {10.5281/zenodo.5143773},
  url          = {https://doi.org/10.5281/zenodo.5143773}
}

@misc{schuhmann2021laion400mopendatasetclipfiltered,
      title={LAION-400M: Open Dataset of CLIP-Filtered 400 Million Image-Text Pairs}, 
      author={Christoph Schuhmann and Richard Vencu and Romain Beaumont and Robert Kaczmarczyk and Clayton Mullis and Aarush Katta and Theo Coombes and Jenia Jitsev and Aran Komatsuzaki},
      year={2021},
      eprint={2111.02114},
      archivePrefix={arXiv},
      primaryClass={cs.CV},
      url={https://arxiv.org/abs/2111.02114}, 
}

@misc{li_andreeto_ranzato_perona_2022, title={Caltech 101}, DOI={10.22002/D1.20086}, abstractNote={Pictures of objects belonging to 101 categories. About 40 to 800 images per category. Most categories have about 50 images. Collected in September 2003 by Fei-Fei Li, Marco Andreetto, and Marc'Aurelio Ranzato. The size of each image is roughly 300 x 200 pixels. We have carefully clicked outlines of each object in these pictures, these are included under the 'Annotations.tar'. There is also a MATLAB script to view the annotations, 'show_annotations.m'.}, publisher={CaltechDATA}, author={Li, Fei-Fei and Andreeto, Marco and Ranzato, Marc'Aurelio and Perona, Pietro}, year={2022}, month={Apr} }

@ARTICLE{6296535,
  author={Deng, Li},
  journal={IEEE Signal Processing Magazine}, 
  title={The MNIST Database of Handwritten Digit Images for Machine Learning Research [Best of the Web]}, 
  year={2012},
  volume={29},
  number={6},
  pages={141-142},
  doi={10.1109/MSP.2012.2211477}}

@InProceedings{parkhi12a,
  author       = "Omkar M. Parkhi and Andrea Vedaldi and Andrew Zisserman and C. V. Jawahar",
  title        = "Cats and Dogs",
  booktitle    = "IEEE Conference on Computer Vision and Pattern Recognition",
  year         = "2012",
}

@article{Safieddine_Kachenoura_Albera_Birot_Karfoul_Pasnicu_Biraben_Wendling_Senhadji_Merlet_2012, title={Removal of muscle artifact from EEG data: Comparison between stochastic (ICA and CCA) and deterministic (EMD and wavelet-based) approaches}, volume={2012}, DOI={10.1186/1687-6180-2012-127}, number={1}, journal={EURASIP Journal on Advances in Signal Processing}, author={Safieddine, Doha and Kachenoura, Amar and Albera, Laurent and Birot, Gwénaël and Karfoul, Ahmad and Pasnicu, Anca and Biraben, Arnaud and Wendling, Fabrice and Senhadji, Lotfi and Merlet, Isabelle}, year={2012}, month={Jul}}

\appendix

\section{Pre-trained Model Details}
\label{sec:app-model-details}

The following table gives more details about the pre-trained vision transformers that serve as the backbone models for our experiments. The training method is either \cite{steiner2022trainvitdataaugmentation} which is denoted as 'Standard' or the method described by \cite{radford2021learningtransferablevisualmodels} which is denoted as 'CLIP.' The source of the models is either the PyTorch Image Models library \cite{rw2019timm}, denoted 'timm', or the OpenCLIP library \cite{ilharco_gabriel_2021_5143773}, denoted 'OpenCLIP.'  

\onecolumn
\begin{table}[h!]
\label{tab:model-details}
\centering
\caption{Further details about the pre-trained vision transformers used in this work.}
\begin{tabular}{|c|c|c|c|c|c|c|}
\hline
 Model & Param. Count (M) & Training Method & Source & Pre-train Dataset & Fine-tune Dataset & Repr. Dim. \\ 
 \hline
 ViT-T & 5.7 & Standard & timm & IN-21k & IN-1k & 192 \\ \hline
 ViT-S & 22.1 & Standard & timm & IN-21k & IN-1k & 384 \\ \hline
 ViT-B & 86.6 & Standard & timm & IN-21k & IN-1k & 768 \\ \hline
 ViT-B & 86.6 & CLIP & OpenCLIP & Laion-400M & N/A & 512 \\ \hline
 ViT-L & 304.3 & Standard & timm & IN-21k & IN-1k & 1024 \\ \hline
 ViT-L & 304.3 & CLIP & OpenCLIP & Laion-400M & N/A & 768 \\ \hline

\end{tabular}
\label{tab:model-details}
\end{table}

\end{document}